\documentclass{article}

\PassOptionsToPackage{numbers}{natbib}
\usepackage[preprint]{neurips_2026}

\usepackage[utf8]{inputenc}
\usepackage[T1]{fontenc}
\usepackage{amsmath}
\usepackage{amsfonts}
\usepackage{microtype}
\usepackage[hyphens]{url}
\usepackage[colorlinks=true, linkcolor=blue, citecolor=blue, urlcolor=blue, breaklinks]{hyperref}
\usepackage{booktabs}
\usepackage{enumitem}
\usepackage{pgfplots}
\pgfplotsset{compat=1.18}

\title{Continued AI Scaling Requires Repeated Efficiency Doublings}
\author{Chien-Ping Lu}
\date{}

\begin{document}

\maketitle

\begin{abstract}
\noindent
This paper argues that continued AI scaling requires repeated efficiency doublings. Classical AI scaling laws remain useful because they make progress predictable despite diminishing returns, but the compute variable in those laws is best read as \emph{logical compute}, not as a record of one fixed physical implementation. Practical burden therefore depends on the efficiency with which physical resources realize that compute. Under that interpretation, diminishing returns mean rising operational burden, not merely a flatter curve. Sustained progress then requires recurrent gains in hardware, algorithms, and systems that keep additional logical compute feasible at acceptable cost. The relevant analogy is Moore's Law, understood less as a theorem than as an organizing expectation of repeated efficiency improvement. AI does not yet have a single agreed cadence for such gains, but recent evidence suggests trends that are at least Moore-like and sometimes faster. The paper's claim is therefore simple: if AI scaling is to remain active, repeated efficiency doublings are not optional. They are required.
\end{abstract}

\section{Introduction}

Classical AI scaling laws, especially for pre-training, are unusually compact. Under compute-optimal conditions, training loss is often well approximated by a power law in compute \cite{kaplan2020scaling, hoffmann2022chinchilla}:
\[
L \propto C^{-\kappa},
\]
where \(L\) denotes loss, \(C\) denotes compute, and \(\kappa\) is the scaling exponent. This is the ordinary effectiveness of scaling laws: they make progress forecastable. Returns decline, but they do not become arbitrary.

But prediction is only the first fact. Scaling laws are also unreasonably effective in a stronger, Wigner-like sense \cite{wigner1960unreasonable}. They remain surprisingly portable across changing architectures, precisions, sparsity patterns, and training-adjacent regimes, and AI progress has continued despite the diminishing returns they encode. The deeper question is therefore not only why the law predicts, but why it keeps traveling and why movement along it keeps happening in practice.

The thesis of this paper is direct. Continued AI scaling requires repeated efficiency doublings. The reason is that the compute variable in the classical law is best read not as a record of a specific implementation, but as \emph{logical compute}: an abstract measure of model-side work. Once that distinction is made, the practical burden of scaling depends on a separate efficiency term that governs how physical resources are converted into logical compute. The same abstraction that makes the law portable also creates a persistent efficiency game in hardware, algorithms, and systems, because it leaves many ways to realize additional logical compute more cheaply.

This framing keeps the paper focused on one interpretive claim and one formal extension. The interpretive claim is that \(C\) in the classical law should be read as logical compute. The formal extension is that once efficiency compounds over time, progress depends not only on the static exponent \(\kappa\), but on the rate of efficiency doubling. Under that view, diminishing returns are not merely a flatter loss curve. They are rising operational burden, and repeated efficiency doublings become a practical requirement for sustaining scaling. The natural historical analogy is Moore's Law, understood not as a theorem of nature but as a durable engineering expectation that organized investment around repeated gains in effective capability \cite{moore1965cramming}.

Recent work extends classical scaling analyses to inference-aware training, precision, sparse routing, and test-time compute allocation \cite{sardana2024beyond,kumar2024precision,ludziejewski2024fgmoe,ludziejewski2025jointmoe,snell2024testtime,chen2024provable}. Those refinements matter, but they are not the paper's main target. Here they serve as evidence that the empirical compute--loss relation remains useful across changing realization choices. The goal is not another fitted variant of the law. It is to clarify what the classical variable means and what the classical formulation leaves out when used as a guide to real progress.

The paper nevertheless remains deliberately centered on pre-training, because that is where the classical compute--loss law is cleanest and best established. The broader hypothesis, however, is more ambitious. Recent work on test-time compute scaling and reinforcement-learning post-training suggests that related scaling regularities may also emerge after pre-training \cite{snell2024testtime,chen2024provable,tan2025rlposttraining,khatri2025scalerl}. If power-law-like relations between performance and effective compute continue to appear in post-training, inference-aware, and test-time settings, then the same basic interpretation should travel as well: practical progress in those regimes should also depend on the efficiency with which effective compute is realized. There is not yet a field-wide consensus on one exact mathematical form for post-training or inference-time scaling, and this paper does not claim to settle that question. It claims only that the same organizing idea is likely to apply beyond pre-training wherever a stable power-law-like relation continues to guide practice.

The paper therefore proceeds in three main moves. First, it restates the classical law and identifies what it omits. Second, it separates logical compute from physical burden by making hidden efficiency explicit. Third, it introduces a time-indexed efficiency-doubling extension that connects static diminishing returns to dynamic, continuing progress. The later sections then draw out the operational and architectural implications of that extension.

\section{Classical Scaling Law and What It Omits}

The standard starting point in the literature is a separable loss model over model size and dataset size:
\[
L(N, D) \;=\; E + A N^{-\alpha} + B D^{-\beta},
\]
where \(N\) is the number of parameters, \(D\) is the number of training tokens, \(E\) is an irreducible loss floor, and \(\alpha,\beta>0\) are empirical exponents. This form captures a familiar idea: increasing either model size or data improves loss, but with diminishing returns.

To connect this expression to compute, one imposes a training-cost constraint. In dense pre-training, total compute is commonly approximated as
\[
C \propto N D,
\]
or, in a standard back-of-the-envelope form,
\[
C \approx 6 N D,
\]
where the constant reflects forward pass, backward pass, and parameter-update costs. Under a fixed compute budget, the compute-optimal choice of \(N\) and \(D\) balances the two loss terms, yielding
\[
N^* \propto C^{\tfrac{\beta}{\alpha+\beta}},
\qquad
D^* \propto C^{\tfrac{\alpha}{\alpha+\beta}}.
\]
Substituting these relations back into the loss gives the familiar compute-only law
\[
L(C) \approx E + K C^{-\kappa},
\qquad
\kappa = \frac{\alpha\beta}{\alpha+\beta},
\]
for some constant \(K>0\).

This derivation helps clarify both the power and the limits of the classical law. It captures a robust empirical relationship between loss and compute, but it does so at a highly abstract level. In particular, \(\kappa\) is a property of how loss responds to scaling in parameters and data under a compute budget. It is not itself a measure of hardware quality, precision format, or system efficiency.

That is why the classical law is best understood as a ratio law. It does not claim that a particular absolute compute budget guarantees a universal level of capability. Rather, it specifies how loss changes \emph{relative} to changes in compute. This is what gives the law its immediate predictive value, and it is also part of what makes scaling laws portable across changing training stacks: they preserve a higher-level relation while abstracting away from many of the mechanisms that realize compute in practice.

Recent extensions that incorporate inference demand, numerical precision, or Mixture-of-Experts structure \cite{sardana2024beyond,kumar2024precision,ludziejewski2024fgmoe,ludziejewski2025jointmoe} do not overturn this basic picture. They show that one can refine the law by exposing variables that the baseline formulation leaves implicit. For the present paper, however, the important point is simpler: the empirical compute--loss relation continues to survive across changing realization choices. That is evidence for the robustness of the abstraction, not a reason to replace the present paper's central question with a survey of every variant. Put differently, pre-training is the clearest anchor case, but the paper's interpretive logic is meant as a broader hypothesis about regimes in which similarly stable power-law-like relations appear, even if their exact mathematical form remains unsettled.

Accordingly, the contribution here is not a new fitted law for a particular architecture or regime. It is interpretive and unifying. The paper proposes a stable reading of the compute variable in classical scaling laws, separates that variable from the efficiency with which it is realized, and then uses that separation to introduce a simple dynamic extension indexed by efficiency doublings over time.

\section{Logical Compute and Hidden Efficiency}

The central omission in the compact law \(L(C) \approx E + K C^{-\kappa}\) is not compute itself, but the \emph{realization} of compute. Classical scaling laws treat compute as an input, yet they do not ask how long it takes to supply that compute, how much power it requires, or how much system engineering is needed to sustain it. In that sense, efficiency is hidden rather than absent.

This motivates a distinction between two quantities:
\begin{itemize}[leftmargin=1.5em]
    \item \textbf{Logical compute}: the abstract model-side work that enters the scaling law, defined against a dense, uniform-precision reference realization of the model rather than against any particular sparse, quantized, or kernel-specific implementation.
    \item \textbf{Physical resource burden}: the time, power, hardware, and systems effort required to realize a given amount of logical compute in practice.
\end{itemize}

We can express their relationship in reduced form by writing the physical resource burden as
\[
PT = \frac{C_{\mathrm{logical}}}{E_{\mathrm{logical}}},
\]
where \(P\) denotes a power budget, \(T\) denotes elapsed time, so \(PT\) is energy, and \(E_{\mathrm{logical}}\) summarizes how efficiently hardware, software, precision, and architecture convert physical resources into logical compute. This makes the core claim explicit: the same logical compute target can impose very different real-world burdens depending on efficiency, and it also makes clear why electricity supply is critical to sustaining continued scaling.

Equivalently, logical compute efficiency can be written as
\[
E_{\mathrm{logical}} = \frac{C_{\mathrm{logical}}}{PT},
\]
which measures how much logical compute is produced per unit of physical resource expenditure. Dimensionally, if \(C_{\mathrm{logical}}\) is measured in logical FLOP-equivalents, \(P\) in watts, and \(T\) in seconds, then \(E_{\mathrm{logical}}\) has units of logical FLOPs per joule.

For present purposes, \(E_{\mathrm{logical}}\) should not be read as one indivisible constant. A useful reduced-form decomposition is
\[
E_{\mathrm{logical}}
\approx
E_{\mathrm{silicon}}
E_{\mathrm{numerical}}
E_{\mathrm{architectural}}
E_{\mathrm{systems}},
\]
where \(E_{\mathrm{silicon}}\) summarizes device-level and memory-energy efficiency, \(E_{\mathrm{numerical}}\) captures gains from lower-precision arithmetic and quantization, \(E_{\mathrm{architectural}}\) captures gains from sparsity, routing, and other changes that reduce the physical cost of delivering a target amount of model-side work, and \(E_{\mathrm{systems}}\) captures kernels, communication overlap, scheduling, and utilization. The factorization is only heuristic, since the terms interact strongly in practice, but it makes the paper's point sharper: an observed efficiency-doubling rate is an aggregate property of the whole stack, not a property of hardware alone.

This distinction is already implicit in engineering practice. Model FLOPs Utilization (MFU) is typically written as
\[
\mathrm{MFU}
=
\frac{\text{model FLOPs}/\text{elapsed time}}{\text{vendor FLOPs/s}}.
\]
If the MFU numerator is identified with logical-compute throughput and the vendor peak is denoted by \(F_{\mathrm{peak}}\), then
\[
\mathrm{MFU} = \frac{C_{\mathrm{logical}}/T}{F_{\mathrm{peak}}}
:= \frac{E_{\mathrm{logical}} P}{F_{\mathrm{peak}}}.
\]
MFU and logical compute efficiency are therefore related but not identical. MFU compares realized model-side throughput against a vendor-defined hardware peak, whereas \(E_{\mathrm{logical}}\) compares logical compute against physical resources such as time and power. In that sense, the industry has already been using something close to logical compute efficiency, but in disguised and peak-normalized form. The numerator is similar in spirit in both cases because it refers to model-implied work. The denominator, however, is very different. Vendor FLOPs/s figures depend on precision conventions, kernel assumptions, and peak reporting conditions, so they are not especially stable as a cross-system baseline. A resource-based denominator is conceptually cleaner for the present argument because it tracks the real burden required to realize a given amount of logical compute.

For that reason, the model-side numerator is the natural candidate for the compute variable in a scaling law. Historically, ``FLOPs'' referred to floating-point operations in a specific numeric format, but in current practice the term often functions more like shorthand for model-side work. Under the interpretation proposed here, the quantity entering the law is better thought of as logical compute: the work implied by the model's dense, uniform-precision forward and backward passes, independent of the particular implementation path used to realize it.

One reason this abstraction is useful is practical rather than merely terminological. Engineers typically do not aim to change the target loss curve arbitrarily. They aim to preserve, as closely as possible, the loss behavior that a dense, uniform-precision reference model would have achieved, while realizing that behavior at lower power, lower cost, or higher throughput through better kernels, sparsity, quantization, routing, or systems design. When those optimizations succeed, they do not overturn the scaling law so much as make it cheaper to remain on it.

The dense, uniform-precision reference should therefore be read as a counterfactual anchor, not as a prediction that frontier systems will remain dense or uniform in precision. Even when the best implementation uses Mixture-of-Experts routing or aggressive formats such as FP8 or INT4, the practical question is still whether those choices preserve or approximate the model-side work that the reference computation would have performed while lowering physical burden. Without some such anchor, every engineering gain risks being absorbed into a redefinition of compute itself, making efficiency harder to distinguish from bookkeeping.

Preserving this distinction is important. If sparsity, quantization, or system optimization are folded into the definition of compute itself, the baseline of the law becomes unstable. The engineering improvements one wants to explain risk being absorbed into a redefinition of the law rather than standing out as efficiency gains. A cleaner interpretation is to keep logical compute fixed and let practical innovations improve \(E_{\mathrm{logical}}\), the efficiency with which it is delivered.

This layered view also helps explain why the burden of scaling has recently shifted. Pure transistor-level improvement no longer carries the whole load. More of the effective doubling rate now has to come from precision choices, memory-system design, routing schemes, custom kernels, communication overlap, and full-stack co-design. That is why recent work on precision, sparse architectures, and systems optimization is best understood as work on \(E_{\mathrm{logical}}\): it raises the amount of logical compute that can be delivered per joule and per unit time even when raw silicon gains alone are insufficient \cite{kumar2024precision,ludziejewski2024fgmoe,ludziejewski2025jointmoe,deepseek2024v3,kimi2026attnres}.

\section{A Time-Indexed Efficiency-Doubling Extension}

Once efficiency is explicit, the next question is temporal: how quickly can logical compute be accumulated over calendar time? Let \(t\) denote elapsed years. In this section, \(\mathcal{E}(t)\) denotes the same logical compute efficiency introduced earlier, now written as a function of time. If efficiency doubles at an annual rate \(\beta\), measured in doublings per year, then
\[
\mathcal{E}(t) = \mathcal{E}_0 2^{\beta t},
\]
where \(\mathcal{E}_0\) is initial efficiency.

The Moore's-Law analogy should be understood carefully. Moore's Law did not derive a permanent universal cadence from first principles; its practical force was as a coordination device for an industry whose underlying semiconductor regime made repeated doublings plausible for a long stretch \cite{moore1965cramming}. The analogous claim here is not that AI obeys one settled efficiency law. It is that scaling creates a similar engineering imperative: once diminishing returns raise the burden of additional progress, the field organizes around recurrent gains in efficiency. Unlike the semiconductor roadmap, however, AI has no single accepted cadence. Different layers of the stack move at different speeds. Even so, some observed trends have recently been at least Moore-like and sometimes faster. Hernandez and Brown estimate a 16-month doubling in algorithmic efficiency for AlexNet-level ImageNet performance between 2012 and 2019 \cite{hernandez2020ai}, while the AI Index 2025 reports that the inference cost for GPT-3.5-level MMLU performance fell by more than 280x between late 2022 and late 2024 \cite{aiindex2025}. The point is not that one number governs all of AI, but that the practical culture of scaling is already organized around repeated efficiency gains.

A simple calibration makes the \(\beta\) parameter operational. If some efficiency frontier improves by a factor \(R\) over an interval of \(\Delta t\) years, the implied annual efficiency-doubling rate is
\[
\beta = \frac{\log_2 R}{\Delta t}.
\]
Under this mapping, a 16-month doubling corresponds to \(\beta \approx 0.75\) doublings per year, while a 280x cost reduction over two years corresponds to \(\beta \approx 4.1\). These quantities are not directly comparable, because the first is an algorithmic-efficiency estimate for a fixed vision target and the second is an application-level inference-cost measure. But they do show how observed historical improvements can be translated into the dynamic parameter of the model.

If the physical resource budget contributed per year is approximately constant at a baseline level \(P_0\), then cumulative logical compute added by time \(t\) is
\[
\Delta C(t) = \int_0^t \mathcal{E}(s) P_0 \, ds
\;=\; \frac{\mathcal{E}_0 P_0}{\beta \ln 2}\left(2^{\beta t}-1\right).
\]
Normalizing the initial condition so that \(C_0 = \mathcal{E}_0 P_0\) yields
\[
C(t) = C_0 + \Delta C(t)
\;=\; C_0\left(1 + \frac{2^{\beta t}-1}{\beta \ln 2}\right).
\]

This expression is the time-indexed counterpart of the paper's main claim. Even when the underlying scaling law is written over logical compute, the amount available by year \(t\) depends on how rapidly the surrounding stack improves its efficiency.

Now combine this with the excess-loss form of the classical scaling law,
\[
L(C) \approx E + K C^{-\kappa}.
\]
Define relative excess loss above the irreducible floor by
\[
X(t) \equiv \frac{L(t)-E}{L(0)-E}.
\]
Substituting \(C(t)\) into the classical law gives
\[
X(t) = \left(1 + \frac{2^{\beta t}-1}{\beta \ln 2}\right)^{-\kappa}.
\]

Figure~\ref{fig:efficiency-doublings} visualizes this relation over calendar time. It fixes \(\kappa\) and compares several annual efficiency-doubling rates \(\beta\), with a Moore-like rate as the baseline reference.

\begin{figure}[t]
\centering
\begin{tikzpicture}
\begin{axis}[
    width=0.82\linewidth,
    height=0.42\textheight,
    xlabel={time \(t\) (years)},
    ylabel={relative excess loss \(X(t)\)},
    xmin=0, xmax=20,
    ymin=0.55, ymax=1.02,
    domain=0:20,
    samples=200,
    legend style={at={(0.97,0.97)}, anchor=north east, draw=none, fill=none},
    tick label style={font=\small},
    label style={font=\small},
    grid=both,
    grid style={line width=.1pt, draw=gray!20},
    major grid style={line width=.2pt, draw=gray!35}
]
\addplot[blue, thick] {(1 + (2^(0.5*x) - 1)/(0.5*ln(2)))^(-0.063)};
\addlegendentry{Moore-like baseline, \(\beta=0.5\)}

\addplot[red, thick] {(1 + (2^(0.25*x) - 1)/(0.25*ln(2)))^(-0.063)};
\addlegendentry{slower doubling, \(\beta=0.25\)}

\addplot[violet!80!black, thick] {(1 + (2^(0.75*x) - 1)/(0.75*ln(2)))^(-0.063)};
\addlegendentry{historical algorithmic reference, \(\beta\approx0.75\)}

\addplot[gray!70!black, thick, dashed] {(1 + x)^(-0.063)};
\addlegendentry{no doubling, \(\beta=0\)}
\end{axis}
\end{tikzpicture}
\caption{Relative excess loss over calendar time for several annual efficiency-doubling rates, holding \(\kappa=0.063\) fixed. The blue curve uses a Moore-like baseline of \(\beta=0.5\) doublings per year, the violet curve shows the historical 16-month algorithmic-efficiency rate from Hernandez and Brown \cite{hernandez2020ai}, and the dashed curve shows the non-doubling case with linear compute growth. Faster annual doubling rates accelerate the accumulation of logical compute and therefore the decline in excess loss.}
\label{fig:efficiency-doublings}
\end{figure}

This equation makes explicit what the static law leaves hidden. The exponent \(\kappa\) still governs the severity of diminishing returns, but progress over calendar time also depends on the annual efficiency-doubling rate \(\beta\). The classical law is static: it tells us how much logical compute is needed. The time-indexed extension is dynamic: it tells us how quickly that logical compute becomes available as efficiency improves year by year. AI scaling therefore depends not only on larger raw budgets, but on sufficiently rapid gains in efficiency to keep the law economically productive. On the timescale of Figure~\ref{fig:efficiency-doublings}, the AI Index inference-cost rate would lie well above the plotted curves, which is exactly the point: different layers of the stack can imply very different effective \(\beta\) values, and the empirical task is to determine which regime is relevant for the question at hand.

\section{The Operational Meaning of Diminishing Returns}

Under the classical reading, diminishing returns mean that further reductions in loss require disproportionately more compute. That statement is mathematically correct, but it becomes much sharper once efficiency is made explicit. If
\[
L(C) \approx E + K C^{-\kappa},
\]
then achieving a target loss \(L_{\mathrm{target}} > E\) requires logical compute on the order of
\[
C_{\mathrm{logical}}(L_{\mathrm{target}})
\propto
(L_{\mathrm{target}}-E)^{-1/\kappa}.
\]
Using \(PT = C_{\mathrm{logical}}/E_{\mathrm{logical}}\), the corresponding physical resource burden is
\[
PT(L_{\mathrm{target}})
\propto
\frac{(L_{\mathrm{target}}-E)^{-1/\kappa}}{E_{\mathrm{logical}}}.
\]

The key point is that diminishing returns are not just a property of curve shape. They are a statement about escalating operational burden. As target loss approaches the irreducible floor, the logical compute requirement grows rapidly because of the power law, and the physical resource burden grows faster still when efficiency improvements fail to keep up.

This perspective also clarifies why progress in AI depends so strongly on systems and architecture. Even if the compute--loss relation remains unchanged, practical progress depends on how effectively the training stack converts real resources into logical compute. Lower precision, better kernels, improved communication patterns, and sparse architectures matter not because they rewrite the classical law, but because they improve the efficiency term that determines whether movement along it remains feasible.

It also suggests a more concrete interpretation of the ``race to efficiency.'' To sustain a roughly steady pace of improvement, the support system behind scaling must itself improve recurrently. That improvement may come partly from larger clusters or higher power budgets, but sustained progress becomes much easier when efficiency compounds as well. In this sense, the power-law form does more than describe diminishing returns; it structures the innovation problem. It reveals why hardware, algorithms, and systems must search for repeated gains in effective efficiency. Systems such as DeepSeek-V3 or Kimi's recent attention optimizations are therefore best understood as responses to scaling pressure rather than exceptions to it \cite{deepseek2024v3,kimi2026attnres}.

\section{Brief Implications}

If scaling laws abstract over realization details, then those details become the margin of competition. Hardware efficiency, kernels, quantization, sparsity, routing, and systems design matter because they improve \(E_{\mathrm{logical}}\), not because each one requires a new theory of progress. The empirical law tells us what additional logical compute buys; the efficiency game changes what that compute costs.

This also clarifies the broader implication of the paper's argument. AI scaling remains empirically active and economically meaningful despite diminishing returns because the law is stable at the level of logical compute, while the burden of realizing that compute is continually renegotiated by the efficiency stack. Hardware, algorithms, and systems remain central not as side notes to scaling, but as the mechanisms that keep movement along the law feasible in practice. On this view, repeated efficiency doublings are not a bonus to scaling. They are what sustained scaling requires.

\section{Conclusion}

Classical AI scaling laws remain powerful because they express a stable ratio relation between loss and compute while abstracting away from many realization details. That abstraction is the source of their unusual usefulness. It is why the law travels across regimes, and why so much engineering effort can be redirected into the efficiency stack without destroying the law's usefulness.

Once the compute variable is interpreted as logical compute, the practical meaning of diminishing returns changes. The central problem is no longer only geometric flattening in a loss curve. It is the rising operational burden of realizing each further unit of useful progress. Continued AI scaling therefore depends jointly on the exponent \(\kappa\) and on the rate at which the surrounding hardware, algorithmic, and systems stack can deliver repeated efficiency doublings.

There is not yet a single agreed cadence for those doublings, and AI should not be described as if it already had one law as crisp as Moore's. But if scaling is to remain active despite diminishing returns, repeated efficiency doublings are required. That is the paper's central claim.

\end{document}